\documentclass[runningheads]{llncs}
\usepackage{graphicx}
\usepackage{tikz}
\usepackage{pgfplots}
\usepgfplotslibrary{statistics}
\usepackage{amsmath,amsfonts}
\usepackage{xcolor}
\usepackage{url}
\usepgfplotslibrary{fillbetween}
\usepackage{floatrow}
\usepackage{array}
\newcolumntype{C}[1]{>{\centering\arraybackslash}p{#1}}
\newfloatcommand{capbtabbox}{table}[][\FBwidth]
\newcommand{\printfnsymbol}[1]{%
    \textsuperscript{\@fnsymbol{#1}}%
}
\newcommand{\figref}[1]{Fig.~\ref{#1}}
\newcommand{\secref}[1]{Section~\ref{#1}}
\newcommand{\tabref}[1]{Table~\ref{#1}}

\usepackage{capt-of}
\pgfplotsset{compat=1.17}
\begin{document}
\title{ADG-Pose: Automated Dataset Generation for Real-World Human Pose Estimation}
\titlerunning{ADG-Pose}
%
\author{Ghazal Alinezhad Noghre\thanks{Authors have equal contribution.} \and Armin Danesh Pazho\printfnsymbol{1} \and Justin Sanchez \and Nathan Hewitt \and Christopher Neff \and Hamed Tabkhi}
\authorrunning{G. Alinezhad Noghre et al.}
%
\institute{University of North Carolina, Charlotte NC 28223, USA}
\maketitle
\begin{abstract}
Recent advancements in computer vision have seen a rise in the prominence of applications using neural networks to understand human poses. However, while accuracy has been steadily increasing on State-of-the-Art datasets, these datasets often do not address the challenges seen in real-world applications. These challenges are dealing with people distant from the camera, people in crowds, and heavily occluded people. As a result, many real-world applications have trained on data that does not reflect the data present in deployment, leading to significant underperformance. This article presents ADG-Pose, a method for automatically generating datasets for real-world human pose estimation. ADG-Pose utilizes top-down pose estimation for extracting human keypoints from unlabeled data. These datasets can be customized to determine person distances, crowdedness, and occlusion distributions. Models trained with our method are able to perform in the presence of these challenges where those trained on other datasets fail. Using ADG-Pose, end-to-end accuracy for real-world skeleton-based action recognition sees a 20\% increase on scenes with moderate distance and occlusion levels, and a 4X increase on distant scenes where other models failed to perform better than random.
    
    \keywords{Human Pose Estimation  \and Real-World \and Data Generation. \vspace{-10pt}}
\end{abstract}

\section{Introduction} \label{sec:Intro}
Human Pose Estimation has seen vast improvements in recent years. This accuracy increase has led to their adoption in real-world applications that benefit from understanding human poses. Smart surveillance, public safety, medical assistance \cite{REVAMP2T,skeletonsurveillance,PatientPose}; are examples of real-world applications that rely on pose information. Unfortunately, despite the current State-of-the-Art (SotA) achieving upwards of 80-90\% accuracy on popular datasets, that accuracy often fails to transfer to real-world scenarios. The number of high-quality datasets with human pose annotations is alarmingly small, as creating them is expensive and time-consuming. Real-world applications are often trained on one of these few datasets, regardless of whether the dataset represents the type of scenes present in deployment.

The disconnect of the training data and inference data (i.e. data seen during deployment) often leads to high-accuracy models, when tested on datasets, underperforming in real-world applications. This disconnect is exceptionally strong in applications that need to detect persons in crowded scenes, heavily occluded persons, or persons very distant from the camera, particularly if the application uses bottom-up pose estimation. A few datasets have been introduced to address some of these issues \cite{CrowdPose,Pose2Seg,tinypeoplepose}, but they all only address a single issue at a time. Further, they use different skeletal structures, making it difficult to utilize them to train a single network. As such, there is a need for datasets that fill the gaps that are left by the current offering.

This article proposes ADG-Pose, a method for generating datasets designed specifically for real-world applications. ADG-Pose allows for the customization of the data distribution along three axes: distance from the camera, crowdedness, and occlusion. ADG-Pose uses high-accuracy models trained on existing datasets to annotate ultra-high resolution images. From there, high-resolution images are created that fit within the distribution parameters set by the user, resulting in a machine annotated dataset customized towards the target real-world application. To validate our method, we create Panda-Pose, a custom dataset suited towards parking lot surveillance. We take a model previously trained on COCO \cite{COCO} and train it on Panda-Pose. We provide comparisons between the two models on both COCO and Panda-Pose, including F1-score to account for false negatives. We also provide qualitative results that show what validation accuracy fails to; models trained on Panda-Pose detect people completely missed by those trained on COCO. Often, these are not even annotated, whether because they are too distant from the camera, in too large a crowd, or too occluded, and do not contribute to validation accuracy.

As a final test of real-world viability, we compare how models trained on Panda-Pose and those trained on COCO affect end-to-end accuracy when used as a backbone for real-world skeleton-based action recognition on the UCF-ARG dataset \cite{ucf-arg}. When using Panda-Pose for training, we see an increase of \textbf{20\%} and \textbf{30\%} on the ground and rooftop scenarios respectfully. For the rooftop scenario, the COCO-trained models resulted in an accuracy equivalent to random guessing.

In summary, this paper encompasses the following contributions:
\begin{enumerate}
    \item We identify and formulate the data gaps and limitations of existing publicly available datasets for real-world human pose estimation.  
    \item We propose ADG-Pose, a novel method for the automated creation of new datasets that address real-world human pose estimation, customizing for distance from camera, crowdedness, and occlusion.\footnote{Code available at https://github.com/TeCSAR-UNCC/ADG-Pose}
    \item We present Panda-Pose, an extension over the existing Panda dataset, to demonstrate the benefits of ADG-Pose to address real-world pose estimation in smart video surveillance applications.
    \item We further demonstrate the benefits of ADG-Pose and Panda-Pose in context of real-world skeleton-based action recognition.
\end{enumerate}

\section{Related Work} \label{sec:RelatedWork}
Keypoint-based human pose estimation can largely be separated into two main categories: top-down methods that work off person crops and bottom-up methods that work off entire scenes. Top-down methods are generally used for single person pose estimation and are assumed to have person crops provided to them \cite{td3,stacked_hour_glass,hrnet_pose}. Top-down methods can be adapted for multi-person pose estimation by attaching them to an assisting detection network that generates person crops \cite{rmpe,td14}. In contrast to top-down methods, bottom-up methods look at the entire scene image and detect all keypoints for all persons at once, using further processing to group them to each individual \cite{openpose2018,pifpaf,personlab,higherhrnet,LOGO-CAP}. Bottom-up methods are often less computationally complex than top-down methods, as top-down methods have to process data for each individual separately, scaling linearly with the number of persons. In contrast, bottom-up approaches have static computation regardless of the number of persons in a scene. This has led to some works focusing on lightweight inference and real-time performance \cite{lightweight_mobv1,EfficientHRNet}.

MPII \cite{MPII} contains 25k images with 40k persons. Images are taken from YouTube videos and have annotations for 16 keypoint skeletons. COCO \cite{COCO} contains over 200k images and 250k person instances. COCO has 17 keypoint pose annotations for over 150k persons and is widely used to train and validate SotA models. AI Challenger \cite{aichallenger} consists of 300k images containing persons labeled with 14 keypoint skeletons. CrowdPose \cite{CrowdPose} attempts to address the lack of crowded scenes in the previous three datasets. Where MPII, AI Challenger, and COCO have distributions that greatly favor scenes with a low number of persons, CrowdPose creates its dataset by sampling from the other three in a way that guarantees a uniform distribution in the crowdedness of the scenes. CrowdPose contains 20k images with 80k persons annotated with AI Challenger style keypoint skeletons. \cite{Pose2Seg} introduces a new benchmark, OCHuman, that focuses on heavily occluded scenes. Maintaining an average IoU of 0.67, OCHuman has 4731 images and 8110 persons annotated with COCO-style keypoint skeletons. Tiny People Pose \cite{tinypeoplepose} consists of 200 images and 585 person instances labeled with modified MPII style keypoint skeletons. The images are focused on persons far from the camera that take consist of very few pixels. The motivation is to address the lack of distant persons in common human pose datasets. Similar focus on distant detection has been seen in object detection \cite{YOLT,SatImgMultiscale}. 
\section{Real-World Pose Estimation Challenges} \label{sec:Motivation}
There are many challenges when using human pose estimation in real-world applications. Take smart surveillance as an example. The types of locations surveillance cameras are placed are widely varied, even for a single system. In a shopping mall cameras will be installed in stores, hallways, food courts, and parking lots. In a store the camera will be closer to people, there will be fewer people in the scene, and occlusions from the merchandise will be common. In hallways and food courts there will be lots of people at medium to long distances to cameras and crowded scenes and occlusions will be prevalent. In parking lots people will often be very far from the camera and often partially occluded by vehicles. Overall, we have identified three main challenges of real-world human pose estimation:

\begin{enumerate}
    \item \textbf{Wide Variety of Distances:} from an algorithmic perspective, this translates to the number of pixels a person takes up in an image. This can also be looked at as the scale of a person compared to the total image resolution.
    \item \textbf{Occlusions:} where a person is partially obscured by a part of the environment or another person.
    \item \textbf{Crowded Scenes:} many real-world applications will require pose estimation in highly crowded locations. In addition to occlusion, a large number of people can make accurately detecting the poses very challenging.
\end{enumerate}

The major limitation in creating a model that can address all these issues is the data used for training. The most popular datasets (MPII \cite{MPII}, AI Challenger \cite{aichallenger}, COCO \cite{COCO}) mostly consist of unoccluded people who are relatively close to the camera in non-crowded scenes. While specialized datasets have been introduced to address some of these concerns (CrowdPose \cite{CrowdPose}, OCHuman \cite{Pose2Seg}, Tiny People Pose \cite{tinypeoplepose}), they each only address a single issue at a time, and their diverse annotation style and validation methods make it challenging to utilize them all for training a single model. Currently, no single dataset can adequately address the three main challenges of real-world human pose estimation.

\begin{figure}[b]
    \centering    \vspace{-15pt}
    \includegraphics[width=\linewidth, trim = 18 18 18 18, clip]{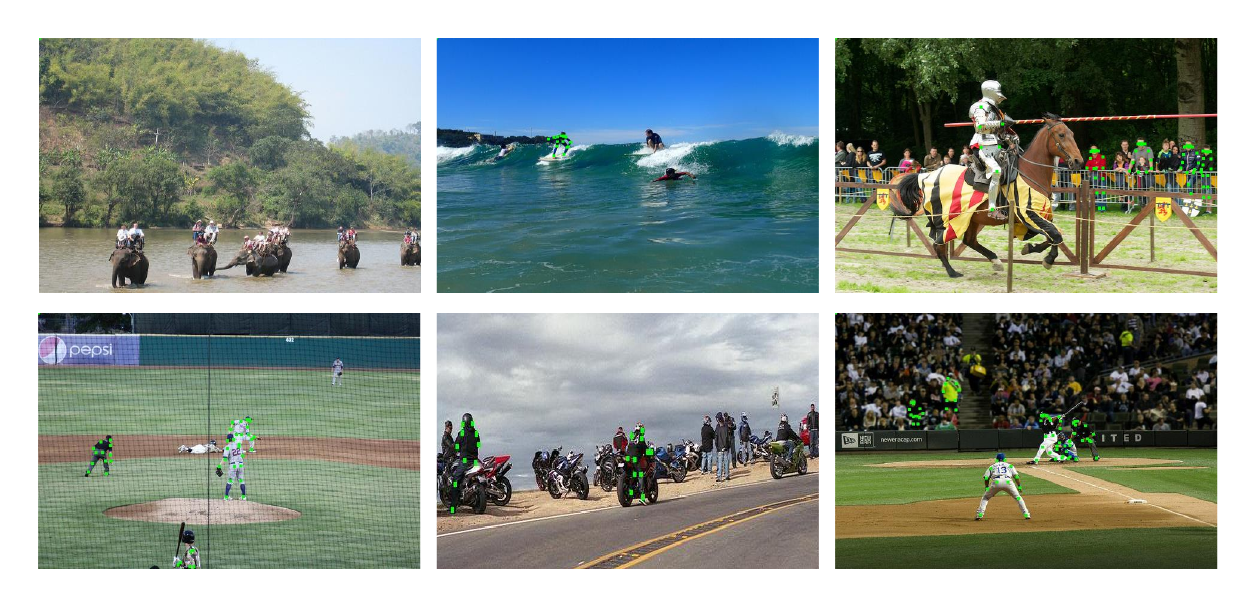}
    \vspace{5pt}
    \caption{Ground truth keypoint annotations (green) from COCO dataset.}
    \label{fig:cocogt}
    \vspace{-15pt}
\end{figure}

\figref{fig:cocogt} displays keypoint annotations from the most prolific keypoint dataset, COCO. Note how distant persons or those in crowded scenes are not annotated. In the upper left image, all the persons riding elephants are unlabeled. On the bottom right image, the vast majority of the crowd is unlabeled. In the remaining images, persons distant from the camera are unlabeled, despite being clearly visible. To be fair, hand annotating all these unlabeled people would be both difficult and time-consuming, so their absence is understandable. COCO's annotation files include a number of all null keypoint annotations to go along with people who might be in the image but are not annotated. During validation, if extra skeletons that don't have annotations are detected, the number of null key points will be subtracted. COCO automatically disregards all but the 20 skeletons with the highest confidence. This is to make sure the networks are not unjustly penalized for estimating skeletons for unlabeled people. Additionally, accuracy on standard datasets is largely reported based on the "Precision" metric, while the "Recall" metric (which includes false negatives) is often ignored. So even if false negatives still occur, they are automatically disregarded by standard validation metrics (i.e. COCO validation).

These limitations can disproportionately affect bottom-up approaches, often preferred for real-world applications due to their much lower computational complexities and much better real-time execution capabilities. In contrast to top-down approaches, bottom-up methods aim to detect persons on their own. The lack of labels can hurt them in both \textbf{training}, where they do not learn to detect distance people, and \textbf{validation}, where they will not be penalized for the large majority of their false negatives (hallucinating persons that are not actually there).  

\section{ADG-Pose} \label{sec:Method}
We propose ADG-Pose, a method of Automated Dataset Generation for real-world human pose estimation. ADG-Pose aims to address all three mentioned challenges in the previous section. ADG-Pose enables users to determine the person scale, crowdedness, and occlusion density distributions of the generated datasets, allowing for training environments that better match the distributions of the target application. 

\begin{figure}[b!]
    \vspace{-15pt}
    \centering
    \includegraphics[width=\linewidth, height=4cm, trim= 15 15 15 65, clip]{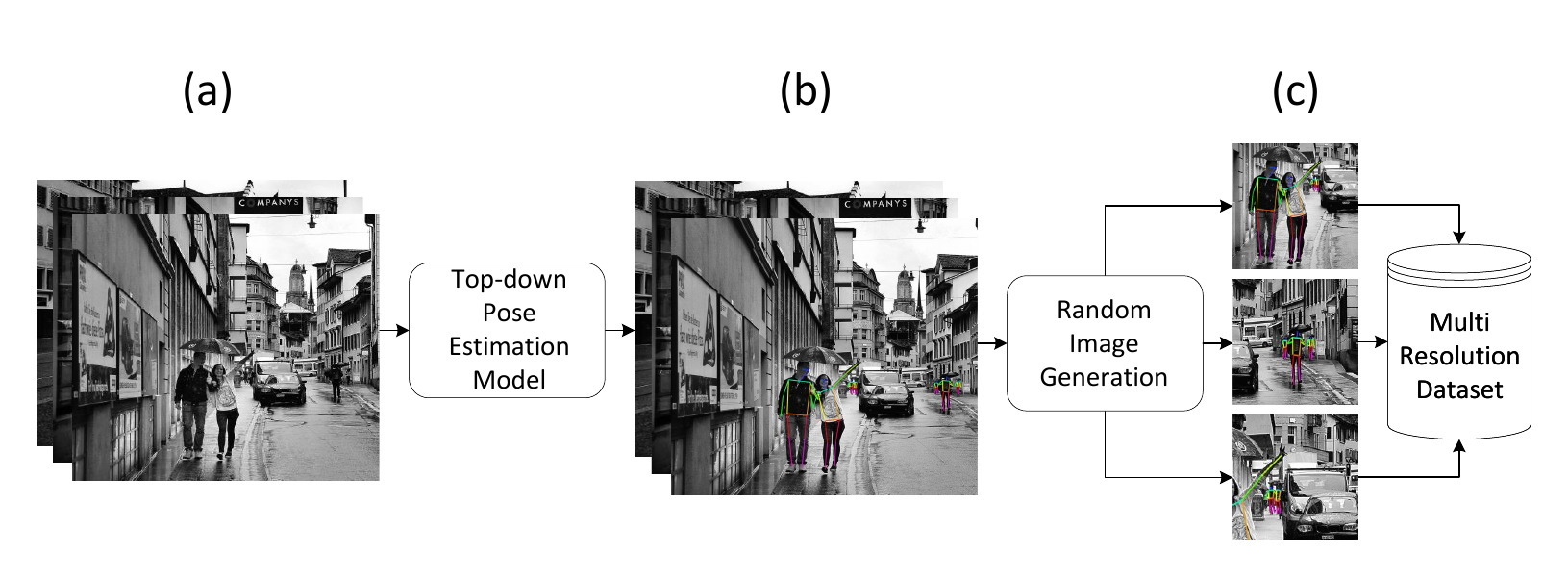}
    \caption{Custom dataset generation. Beginning with ultra high resolution images, a pretrained top-down pose estimation model is used to generate high accuracy keypoint annotations. Semi-random cropping is used to generate numerous high resolution images inline with user specified statistic, forming the new dataset.}
    \label{fig:process}
    \vspace{-5pt}
\end{figure}

\figref{fig:process} shows the three main stages of ADG-Pose. First, a high accuracy top-down human pose estimation model is used to label ultra-high resolution images. By utilizing ultra-high resolution images and a top-down approach, we can mitigate potential issues with annotating distant people as the absolute resolution of the person crops will still be large enough to ensure high accuracy. Second, we take the fully annotated ultra-high resolution images and generate semi-random crops from them. These crops are semi-random because we introduce user-defined parameters to ensure the final dataset will match the desired statistics. First, the user can determine the resolution range to take crops at. To better detect distant persons, larger resolution crops can be used and downscaled to the desired input resolution, thus mimicking larger distances. Second, the user can determine the maximum, minimum, and mean number of persons in a crop. This allows for customization of how crowded a scene is. Third, the user can specify the desired average IoU between people in the crop, tweaking the overall level of occlusion in the dataset. After these crops are made and the statistics verified, the resulting images and annotations are synthesized into a new multi-resolution dataset. Additional user-defined parameters include the total size of the dataset, "train/val/test splits", image aspect ratio, and skeleton/validation style, which must be compatible with the top-down model used for annotation.

\textbf{Panda-Pose:} As a use-case, we choose a real-world application of smart security in an outdoor parking lot environment. For skeleton and validation style, we choose COCO \cite{COCO} as it is currently the most prominent in the field. We use HRNet-w32 \cite{hrnet_pose} for our top-down annotation model and choose PANDA \cite{panda} as our base dataset. PANDA is a gigapixel-level human-centric dataset with an extremely wide field of view ($\sim$1 km\textsuperscript{2}). It contains bounding box annotations for persons, with some scenes containing up to 4k persons. There are 555 frames across four outdoor scenes, which would normally be far too few for training. However, high density and extreme resolution (25k$\times$14k) result in significant information per scene and more than adequate generated images. Additionally, the wide variance in poses, scales, and occlusions allows us to create a range of challenging datasets for different user specifications. We call the resulting dataset Panda-Pose.

For the first specification, parking lots will likely include people quite distant from the camera, resulting in a very small scale. As such, Panda-Pose targets a person scale distribution on the smaller side. Since detecting small-scale persons is more complicated than a large-scale person, we heavily weigh the lower end of the scale spectrum, as can be seen in \figref{fig:b&w}. The wide field of view of most parking lot security cameras will allow for a fair amount of people in the scene, though there is usually enough space that occlusions, while present, will be less than that of more crowded indoor scenes. As such, we target a relatively high number of people per image ($\sim$9) and a moderate amount of occlusions ($\sim$0.33). We also set a maximum of 30 people per image (for fair comparisons in \secref{sec:Results}). Our aspect ratio is 4:3, the maximum resolution is 3840$\times$2880, and the minimum is 480$\times$360. There are 83k training, and 21k validation images with 775k and 202k annotated skeletons, respectively. 4\% of images are without annotations. Training and validation splits have matching distributions.

\begin{figure}[th!]
\CenterFloatBoxes
    \begin{floatrow}
    \ffigbox{%
        \begin{tikzpicture} 
            \begin{axis}
                [
                height=0.8\linewidth,
                width=0.9\linewidth,
                ytick={1,2,3,4},
                yticklabels={\textbf{Panda-Pose}, CrowdPose, OCHuman, COCO},
                xtick={0,0.25,0.5,0.75,1},
                ytick style={draw=none},
                ]
                \addplot[
                boxplot prepared={
                  median=0.216,
                  upper quartile=0.373,
                  lower quartile=0.126,
                  upper whisker=1.0,
                  lower whisker=0.007
                }, draw=black, fill=red,
                ] coordinates {};
                \addplot[
                boxplot prepared={
                  median=0.485,
                  upper quartile=0.773,
                  lower quartile=0.226,
                  upper whisker=1.0,
                  lower whisker=0.008
                },draw=black, fill=orange,
                ] coordinates {};
                \addplot[
                boxplot prepared={
                  median=0.844,
                  upper quartile=0.912,
                  lower quartile=0.752,
                  upper whisker=1.0,
                  lower whisker=0.130
                },draw=black, fill=green,
                ] coordinates {};
                \addplot[
                boxplot prepared={
                  median=0.364,
                  upper quartile=0.613,
                  lower quartile=0.215,
                  upper whisker=1.0,
                  lower whisker=0.034
                },draw=black, fill=cyan,
                ] coordinates {};
            \end{axis}
        \end{tikzpicture}
    }{%
        \caption{Person scale distributions across datasets.}\label{fig:b&w}%
    }
    \killfloatstyle
    \ttabbox
    {\begin{tabular}{c|c|c}
                                        & Persons   & Average \\
                    Dataset             & per Image & IoU     \\
                    \hline \hline                                                                              
                    MPII                & 1.6       & 0.11    \\
                    AI Challanger       & 2.33      & 0.12    \\
                    COCO                & 1.25      & 0.11    \\
                    OCHuman             & 1.72      & 0.67    \\
                    CrowdPose           & 4         & 0.27    \\
                    Tiny People Pose    & 2.93      & -       \\
                    \textbf{Panda-Pose}       & \textbf{9.33} & \textbf{0.33}  \\
                \end{tabular}%
            }    
        {%
            \caption{Person density and occlusion (IoU) across datasets.}\label{tab:stats}%
        }
    \end{floatrow}
\end{figure}

\figref{fig:b&w} and \tabref{tab:stats} present the statistics of Panda-Pose compared to existing popular datasets. Note: stats for Tiny People Pose could not be gathered because the dataset is not publicly available. Overall, the scale distribution in Panda-Pose leans noticeably smaller than other datasets. The closest is COCO, whose scale is about 1.7X larger at every quartile and whose minimum is 5X larger. Additionally, COCO's persons per image and average IoU are significantly lower than Panda-Pose (7.5X and 3X, respectively), putting it well outside our desired statistic. Looking at average IoU, CrowdPose \cite{CrowdPose} comes close enough to seem a suitable replacement. However, CrowdPose has $\frac{1}{2}$ the number of persons per image, and their scale distribution is even worse than COCO's for our application. OCHuman \cite{Pose2Seg} has twice the average IoU, making it far more occluded than Panda-Pose. This could be argued to be a benefit, as detecting with occlusions is significantly more challenging. However, people in OCHuman are generally very close to the camera, taking up nearly the whole image with an average scale of 0.844. All this shows that while other datasets can address part of the challenges for our chosen application, only Panda-Pose addresses them all, matching the desired statics for training and validation.
\section{Results and Evaluation} \label{sec:Results}
To validate the efficacy of ADG-Pose, we train a bottom-up pose estimator on Panda-Pose (\secref{sec:Method}) and use it to compare Panda-Pose with the baseline COCO \cite{COCO} dataset. For the bottom-up pose estimator, we use EfficientHRNet \cite{EfficientHRNet} for its lightweight and real-time execution capabilities, making it more suitable for real-world applications. In addition, its scalability allows us to test with different network complexities. In this article, we use EfficientHRNet's H$_0$ and H$_1$ models. EfficientHRNet by default limits the number of detections to 30, fitting with the COCO dataset. To more fairly compare, we take the same approach when training and validating with our dataset. Training on Panda-Pose starts with pretrained models and is fine-tuned for 150 epochs with a learning rate of $1e-5$ for H$_0$ and $1e-6$ for the larger H$_1$. 

\begin{table}[!bt]
    \centering
        \begin{tabular}{c|c|c|c|c|c}
            Method         & Backbone        & Input Size & AP & AR & F1  \\
            \hline \hline                                                                              
            \multicolumn{6}{c}{trained on COCO}  \\
            \hline                                          
            OpenPose \cite{openpose2018}       & -             & -     & 61.0 & - & - \\
            Hourglass \cite{stacked_hour_glass}      & Hourglass     & 512   & 56.6 & - & - \\
            PersonLab \cite{personlab}      & ResNet-152    & 1401  & 66.5 & - & - \\
            PifPaf \cite{pifpaf}         & ResNet-152    & -     & 67.4 & - & - \\
            HigherHRNet \cite{higherhrnet}    & HRNet-W32     & 512   & 67.1 & - & - \\
            HigherHRNet \cite{higherhrnet}    & HRNet-W48     & 640   & 69.8 & - & - \\
            LOGO-CAP \cite{LOGO-CAP}         & HRNet-W32    & 512   & 69.6 & - & - \\
            LOGO-CAP \cite{LOGO-CAP}         & HRNet-W48    & 640   & 72.2 & - & - \\
            EfficientHRNet-H$_{0}$ \cite{EfficientHRNet} & EfficientNet-B0  & 512   & 64.8 & 69.6 & 67.1 \\
            EfficientHRNet-H$_{1}$ \cite{EfficientHRNet} & EfficientNet-B1  & 544   & 66.3 & 70.7 & 68.4 \\
            \hline 
            \multicolumn{6}{c}{trained on Panda-Pose} \\
            \hline
            EfficientHRNet-H$_{0}$ \cite{EfficientHRNet} & EfficientNet-B0  & 512   & 50.6 & 59.2 & 54.6 \\
            EfficientHRNet-H$_{1}$ \cite{EfficientHRNet} & EfficientNet-B1  & 512   & 48.9 & 56.8 & 52.6 \\
        \end{tabular}
        \vspace{-15pt}
    \caption{Precision, Recall, and F1-score on COCO val.}
    \label{tab:cocoval}
\end{table}

\begin{figure}[t!]
    \centering
    \includegraphics[width=\linewidth, trim=70 18 18 18, clip]{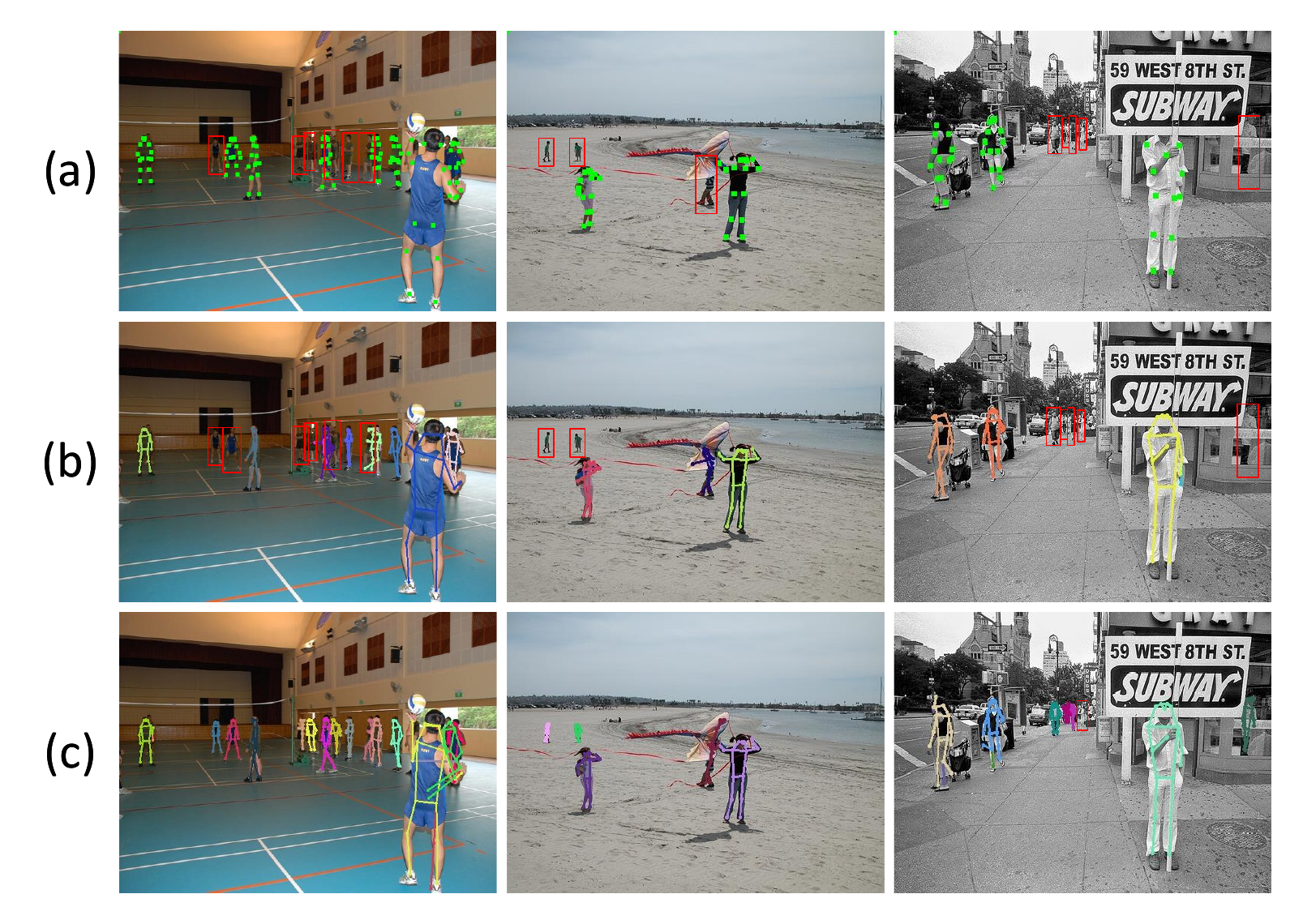}
    \caption{Top: COCO ground truth keypoints (green). Middle: H$_0$ predictions when trained on COCO. Bottom: H$_0$ predictions when trained on Panda-Pose. In all cases, red boxes denote unannotated or undetected persons.} 
    \label{fig:cocoqual}

\end{figure}

\textbf{Evaluation on COCO:} To show how H$_0$ trained on Panda-Pose compares with SotA models trained on COCO, we conduct validation on the COCO dataset. \tabref{tab:cocoval} contains accuracy when validated on COCO val (including precision, recall, and F1-score) while \figref{fig:cocoqual} shows qualitative examples from validation. Looking at the reported validation accuracy, the Panda-Pose trained H$_0$ performs significantly worse than all other models. However, when looking at actual examples from the validation set, we see a completely different story. As discussed in \secref{sec:Motivation} ground truth annotations are missing from distant people or in crowded scenes. This leads to lots of missed detections from COCO trained models, as seen in the center row. Multiple persons in the crowded scene on the left and distant people in the middle and right image are not detected. The Panda-Pose model is able to detect all persons in the first two images and only misses the single most distant person in the last image. However, since these people are not annotated on the COCO dataset, the COCO model does not get penalized for missing them and the Panda-Pose model does not benefit from being able to detect them, at least as far as COCO validation is concerned. However, real-world applications like our test case would weigh being able to detect distant persons much higher. Additionally, while the Panda-Pose model is not perfect, it also attempts to detect highly occluded persons. Looking at the leftmost image, the network greatly misinterprets that person's pose by trying to predict key points for the highly occluded person behind the man serving the ball. Meanwhile, the COCO model does not even detect that person. Another thing to note is how the H$_1$ Panda-Pose model with an input resolution of 768 actually performed worse than H$_0$ on COCO val. This is caused by lower resolution COCO images' upscaling to fit the higher input resolution, leading to additional noise. This is in line with the conclusions made in \cite{higherhrnet}.


\begin{table}[!t]
    \centering
        \begin{tabular}{C{3.4cm}|C{2.9cm}|C{2cm}|C{1cm}|C{1cm}|C{1cm}} 
            Method         & Backbone        & Input Size & AP & AR & F1 \\ 
            \hline
            \hline
            \multicolumn{6}{c}{trained on COCO}    \\
            \hline                                          
            EfficientHRNet-H$_{0}$  & EfficientNet-B0  & 512   & 20.2 & 24.0 & 21.9  \\ 
            EfficientHRNet-H$_{1}$  & EfficientNet-B1  & 544   & 21.1 & 25.1 & 23.4  \\ 
            \hline
            \multicolumn{6}{c}{trained on Panda-Pose} \\                   
            \hline
            EfficientHRNet-H$_{0}$ & EfficientNet-B0  & 512   & 31.4 & 38.7 & 34.7  \\ 
            EfficientHRNet-H$_{1}$ & EfficientNet-B1  & 512   & 34.6 & 44.0 & 38.7  \\ 
            EfficientHRNet-H$_{0}$ & EfficientNet-B0  & 768   & 36.5 & 44.0 & 39.9  \\ 
            EfficientHRNet-H$_{1}$ & EfficientNet-B1  & 768   & 41.3 & 49.9 & 45.2  \\ 

        \end{tabular}
    \vspace{-15pt}
    \caption{Precision, Recall, and F1-score of EfficientHRNet models on Panda-Pose. \vspace{-15pt}}
    \label{tab:ourval}
\end{table}

\textbf{Evaluation on Panda-Pose:} As explored in \secref{sec:Motivation}, the COCO dataset does not accurately represent our target real-world application. Since Panda-Pose was created to closely match our target application we look at how the performance of models trained on COCO compare with those trained on Panda-Pose. This dataset is significantly more challenging than COCO, with \textbf{7.5$\times$} the number of persons per image, \textbf{3$\times$} the occlusions, and a significant shift in distribution towards smaller scale persons. As seen in \tabref{tab:ourval}, EfficientHRNet-H$_0$ trained on COCO barely reaches past \textbf{20\%} AP and has an F1-score of \textbf{21.9\%}. Moving to the H$_1$ model increases AP to \textbf{21.1\%} and F1 to \textbf{23.4\%}. In contrast H$_0$ trained on Panda-Pose reaches an AP of \textbf{31.4\%} and F1 of \textbf{34.7\%}, and increase of \textbf{1.5$\times$} and \textbf{1.6$\times$} respectively. Increasing the resolution of H$_0$ to 768 increases AP to \textbf{36.5\%} and F1 to \textbf{39.9\%}, which is a \textbf{15\%} increase with no other changes to the model. Notably, increases to 768 resolution have a negative effect on COCO accuracy \cite{higherhrnet}, but since Panda-Pose is much higher resolution, performance is improved. This effect is even more prominent than simply increasing the model size without changing the resolution. However, changing the model size to H$_1$ and the resolution to 768, we see an AP of \textbf{41.3\%} and F1 of \textbf{45.2\%}, double what was achievable with the COCO model.

While these results are important, the COCO trained models are at an obvious disadvantage having not trained on Panda-Pose. In addition to the clear challenges of scale, crowdedness, and occlusions, COCO models must combat general domain shift that Panda-Pose models do not. As such, we must use a third dataset unseen by both COCO and Panda-Pose models.

\begin{figure}[t!]
    \centering
    \includegraphics[width=\linewidth, trim=18 18 18 18, clip]{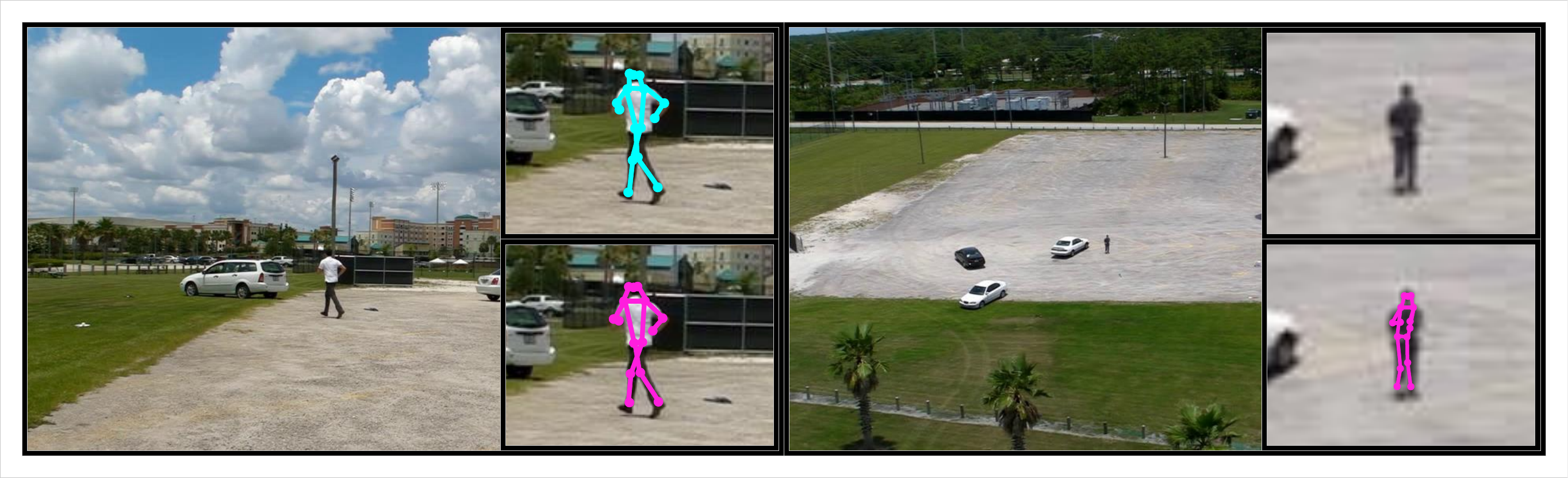}
    \caption{Sample images from UCF-ARG Ground (left) and Rooftop (right) with COCO (blue) and Panda-Pose (pink) predictions. \vspace{-10pt} }
    \label{fig:ucfqual}
\end{figure}

\textbf{Case Study - Action Recognition:} Continuing with the use case of parking lot surveillance, we assess the end-to-end performance of real-world action recognition on the UCF-ARG dataset \cite{ucf-arg}. UCF-ARG consists of 10 actions by 12 actors on three different high resolution (1920$\times$1080) cameras. We focus on the "Ground" and "Rooftop" cameras, as the aerial camera does not fit our use case. We utilize a spatial-temporal graph convolutional network which uses a graph-based formulation to construct dynamic skeletons \cite{ST-GCN}, and add attentive feedback to predict actions, as in \cite{RW-GCN}. The skeletal poses come from H$_0$.

The COCO trained model achieves \textbf{60\%} accuracy on Ground and \textbf{10\%} accuracy on Rooftop, the latter of which is random guessing. As seen in \figref{fig:ucfqual}, the COCO trained model is completely unable to detect the highly distant persons in Rooftop. The model trained on Panda-Pose is able to achieve much better results of \textbf{81\%} on Ground and \textbf{40\%} on Rooftop. Not only is it able to detect more persons in Ground, leading to a \textbf{1.35$\times$} increase in end-to-end accuracy, but it can effectively detect people in Rooftop where the COCO model failed. The significantly smaller person scale distribution of Panda-Pose gives models the ability to accurately detect people much farther from the camera than other datasets, which is an ability completely overlooked in COCO's validation. However, the quality of the poses does slightly suffer from lack of information of very distant persons, as can be seen in \figref{fig:ucfqual}. These results emphasize the efficacy of Panda-Pose and ADG-Pose for real-world applications.


\section{Conclusion} \label{sec:Conclusion}

In this article we presented ADG-Pose for generating datasets for real-world human pose estimation. Current SotA datasets do not always address the challenges faced by real-world applications, which often leads to unexpected under performance. By using ultra-high resolution images and high accuracy neural networks, ADG-Pose allows users to customize datasets towards their chosen application by determining the data distribution along the axes of crowdedness, occlusion, and distance from the camera. We have shown through quantitative and qualitative analysis how validation on current SotA datasets can fail to properly address the challenges of real-world applications, and we have provided real-world skeleton based action recognition as a use case to show how our method produces models better suited for real-world applications.

\textit{Acknowledgements}  This research is supported by the National Science Foundation (NSF) under Award No. 1831795 and NSF Graduate Research Fellowship Award No. 1848727.

\vspace{-10pt}
\bibliographystyle{splncs04}
\bibliography{ref.bib}

\begin{thebibliography}{10}
\providecommand{\url}[1]{\texttt{#1}}
\providecommand{\urlprefix}{URL }
\providecommand{\doi}[1]{https://doi.org/#1}

\bibitem{MPII}
Andriluka, M., Pishchulin, L., Gehler, P., Schiele, B.: 2d human pose
  estimation: New benchmark and state of the art analysis. In: IEEE Conference
  on Computer Vision and Pattern Recognition (CVPR) (June 2014)

\bibitem{openpose2018}
Cao, Z., Hidalgo, G., Simon, T., Wei, S., Sheikh, Y.: Openpose: Realtime
  multi-person 2d pose estimation using part affinity fields. CoRR
  \textbf{abs/1812.08008} (2018), \url{http://arxiv.org/abs/1812.08008}

\bibitem{PatientPose}
{Chen}, K., {Gabriel}, P., {Alasfour}, A., {Gong}, C., {Doyle}, W.K.,
  {Devinsky}, O., {Friedman}, D., {Dugan}, P., {Melloni}, L., {Thesen}, T.,
  {Gonda}, D., {Sattar}, S., {Wang}, S., {Gilja}, V.: Patient-specific pose
  estimation in clinical environments. IEEE Journal of Translational
  Engineering in Health and Medicine  \textbf{6},  1--11 (2018).
  \doi{10.1109/JTEHM.2018.2875464}

\bibitem{td14}
Chen, Y., Wang, Z., Peng, Y., Zhang, Z., Yu, G., Sun, J.: Cascaded pyramid
  network for multi-person pose estimation. CoRR  \textbf{abs/1711.07319}
  (2017), \url{http://arxiv.org/abs/1711.07319}

\bibitem{higherhrnet}
Cheng, B., Xiao, B., Wang, J., Shi, H., Huang, T.S., Zhang, L.: Higherhrnet:
  Scale-aware representation learning for bottom-up human pose estimation
  (2019)

\bibitem{ucf-arg}
for Research~in Computer~Vision, U.C.: Ucf-arg dataset,
  \url{https://www.crcv.ucf.edu/data/UCF-ARG.php}

\bibitem{YOLT}
Etten, A.V.: You only look twice: Rapid multi-scale object detection in
  satellite imagery (2018)

\bibitem{rmpe}
{Fang}, H., {Xie}, S., {Tai}, Y., {Lu}, C.: Rmpe: Regional multi-person pose
  estimation. In: 2017 IEEE International Conference on Computer Vision (ICCV).
  pp. 2353--2362 (2017)

\bibitem{td3}
{Johnson}, S., {Everingham}, M.: Learning effective human pose estimation from
  inaccurate annotation. In: CVPR 2011. pp. 1465--1472 (2011)

\bibitem{pifpaf}
Kreiss, S., Bertoni, L., Alahi, A.: Pifpaf: Composite fields for human pose
  estimation. CoRR  \textbf{abs/1903.06593} (2019),
  \url{http://arxiv.org/abs/1903.06593}

\bibitem{skeletonsurveillance}
Kumar, D., T, P., Murugesh, A., Kafle, V.P.: Visual action recognition using
  deep learning in video surveillance systems. In: 2020 ITU Kaleidoscope:
  Industry-Driven Digital Transformation (ITU K). pp.~1--8 (2020).
  \doi{10.23919/ITUK50268.2020.9303222}

\bibitem{CrowdPose}
Li, J., Wang, C., Zhu, H., Mao, Y., Fang, H.S., Lu, C.: Crowdpose: Efficient
  crowded scenes pose estimation and a new benchmark. In: Proceedings of the
  IEEE/CVF Conference on Computer Vision and Pattern Recognition (CVPR) (June
  2019)

\bibitem{COCO}
Lin, T.Y., Maire, M., Belongie, S., Bourdev, L., Girshick, R., Hays, J.,
  Perona, P., Ramanan, D., Zitnick, C.L., Dollár, P.: Microsoft coco: Common
  objects in context (2014)

\bibitem{REVAMP2T}
{Neff}, C., {Mendieta}, M., {Mohan}, S., {Baharani}, M., {Rogers}, S.,
  {Tabkhi}, H.: Revamp2t: Real-time edge video analytics for multicamera
  privacy-aware pedestrian tracking. IEEE Internet of Things Journal
  \textbf{7}(4),  2591--2602 (2020). \doi{10.1109/JIOT.2019.2954804}

\bibitem{EfficientHRNet}
Neff, C., Sheth, A., Furgurson, S., Tabkhi, H.: Efficienthrnet: Efficient
  scaling for lightweight high-resolution multi-person pose estimation (2021).
  \doi{10.1007/s11554-021-01132-9}

\bibitem{tinypeoplepose}
Neumann, L., Vedaldi, A.: Tiny people pose. In: Asian Conference on Computer
  Vision (ACCV). pp. 558--574. Springer International Publishing (2018).
  \doi{10.1007/978-3-030-20893-6-35}

\bibitem{stacked_hour_glass}
Newell, A., Yang, K., Deng, J.: Stacked hourglass networks for human pose
  estimation. CoRR  \textbf{abs/1603.06937} (2016),
  \url{http://arxiv.org/abs/1603.06937}

\bibitem{lightweight_mobv1}
Osokin, D.: Real-time 2d multi-person pose estimation on {CPU:} lightweight
  openpose. CoRR  \textbf{abs/1811.12004} (2018),
  \url{http://arxiv.org/abs/1811.12004}

\bibitem{personlab}
Papandreou, G., Zhu, T., Chen, L., Gidaris, S., Tompson, J., Murphy, K.:
  Personlab: Person pose estimation and instance segmentation with a bottom-up,
  part-based, geometric embedding model. CoRR  \textbf{abs/1803.08225} (2018),
  \url{http://arxiv.org/abs/1803.08225}

\bibitem{RW-GCN}
Sanchez, J., Neff, C., Tabkhi, H.: Real-world graph convolution networks
  (rw-gcns) for action recognition in smart video surveillance. In: Symposium
  on Edge Computing (SEC). pp. 121--134. ACM/IEEE (Dec 2021).
  \doi{10.1145/3453142.3491293}

\bibitem{hrnet_pose}
Sun, K., Xiao, B., Liu, D., Wang, J.: Deep high-resolution representation
  learning for human pose estimation. CoRR  \textbf{abs/1902.09212} (2019),
  \url{http://arxiv.org/abs/1902.09212}

\bibitem{SatImgMultiscale}
Van~Etten, A.: Satellite imagery multiscale rapid detection with windowed
  networks. 2019 IEEE Winter Conference on Applications of Computer Vision
  (WACV)  (Jan 2019). \doi{10.1109/wacv.2019.00083},
  \url{http://dx.doi.org/10.1109/WACV.2019.00083}

\bibitem{panda}
Wang, X., Zhang, X., Zhu, Y., Guo, Y., Yuan, X., Xiang, L., Wang, Z., Ding, G.,
  Brady, D.J., Dai, Q., Fang, L.: Panda: A gigapixel-level human-centric video
  dataset. In: Computer Vision and Pattern Recognition (CVPR), 2020 IEEE
  International Conference on. IEEE (2020)

\bibitem{aichallenger}
Wu, J., Zheng, H., Zhao, B., Li, Y., Yan, B., Liang, R., Wang, W., Zhou, S.,
  Lin, G., Fu, Y., et~al.: Large-scale datasets for going deeper in image
  understanding. 2019 IEEE International Conference on Multimedia and Expo
  (ICME)  (Jul 2019). \doi{10.1109/icme.2019.00256},
  \url{http://dx.doi.org/10.1109/ICME.2019.00256}

\bibitem{LOGO-CAP}
Xue, N., Wu, T., Zhang, Z., Xia, G.S.: Learning local-global contextual
  adaptation for fully end-to-end bottom-up human pose estimation (2021)

\bibitem{ST-GCN}
Yan, S., Xiong, Y., Lin, D.: Spatial temporal graph convolutional networks for
  skeleton-based action recognition. In: Thirty-second AAAI conference on
  artificial intelligence (2018)

\bibitem{Pose2Seg}
Zhang, S.H., Li, R., Dong, X., Rosin, P., Cai, Z., Han, X., Yang, D., Huang,
  H., Hu, S.M.: Pose2seg: Detection free human instance segmentation. In:
  Proceedings of the IEEE/CVF Conference on Computer Vision and Pattern
  Recognition (CVPR) (June 2019)

\end{thebibliography}
\end{document}